\documentclass{article} 
\usepackage{iclr2026_conference,times}
\usepackage{graphicx}
\usepackage{booktabs}
\usepackage{multirow}

\usepackage{amsmath,amsfonts,bm}

\def\eqref#1{equation~\ref{#1}}

\def\1{\bm{1}}

\DeclareMathAlphabet{\mathsfit}{\encodingdefault}{\sfdefault}{m}{sl}
\SetMathAlphabet{\mathsfit}{bold}{\encodingdefault}{\sfdefault}{bx}{n}

\usepackage{float}
\usepackage{hyperref}
\usepackage{url}

\title{CLIC: Contextual Language-Informed Cardiac Pathology Classification}

\author{Giovani D. Lucafó, Rafael da Costa Silva, João Lucas Luz Lima Sarcinelli,\\ \textbf{Andre Guarnier De Mitri \& Diego Furtado Silva} \\
Institute of Mathematical and Computer Sciences\\
Universidade de São Paulo\\
São Carlos, São Paulo, Brazil \\
\texttt{\{giovanilucafo, rafael.csilva, andremitri,}\\
\texttt{joao.luz,diegofsilva\}@usp.br}
}

\iclrfinalcopy

\track{Research}

\begin{document}

\maketitle

\begin{abstract}
The electrocardiogram (ECG) is the gold standard for non-invasive diagnosis of cardiac pathologies and is a fundamental pillar of cardiovascular medicine. Recent progress in deep learning has led to the development of robust automated classifiers that achieve high performance by processing raw physiological signals. However, in clinical practice, diagnosis is rarely based solely on the signal. Cardiologists commonly support their interpretation with the patient's characteristics and the specific data-acquisition context. Despite this, most current algorithms remain restricted to signal-only analysis, failing to integrate technical metadata and demographic variables. This paper proposes \textbf{C}ontextual \textbf{L}anguage-\textbf{I}nformed \textbf{C}ardiac pathology classification (CLIC), a multimodal framework that significantly enhances diagnostic precision by encoding these variables through natural language. We demonstrate that translating patient-level contextual data into descriptive text provides an informative anchor that helps the model disambiguate complex physiological patterns. We further investigate the use of Large Language Models to synthesize richer clinical descriptions and observe that, while these generated texts remain competitive, controlled template-based contextual clinical text leads to consistent improvements in downstream classification performance.

\end{abstract}

\section{Introduction}

Modern machine learning-based applications increasingly rely on large, heterogeneous datasets that collect information from multiple sources and modalities. In this context, multimodal learning has emerged as an approach to building richer, more complex representations~\citep{li2025}. However,  compared to domains such as vision and language, the integration of different modalities with health time series remains less explored~\citep{trirat2024}. Few studies address the fusion of signals, clinical notes, and structured records in a unified framework~\citep{LIU2026131364, li2023frozenlanguagemodelhelps}.

In healthcare, where time series are widely used to monitor patients, the limitations of unimodal approaches are evident. Medical decision-making relies on multiple sources of information, including physiological signals, demographic data, and textual reports, which provide complementary views of patient health. Relying on a single modality can lead to incomplete or suboptimal diagnosis, whereas richer representations can improve classification performance. Despite the success of deep learning for physiological time series, most existing architectures focus only on signal data and do not leverage the multimodal nature of medical data~\citep{liu2024timeseriesrepresentationlearning, wang2023contrasteverythinghierarchicalcontrastive}.

\begin{figure}[H]
\begin{center}
\includegraphics[width=\linewidth]{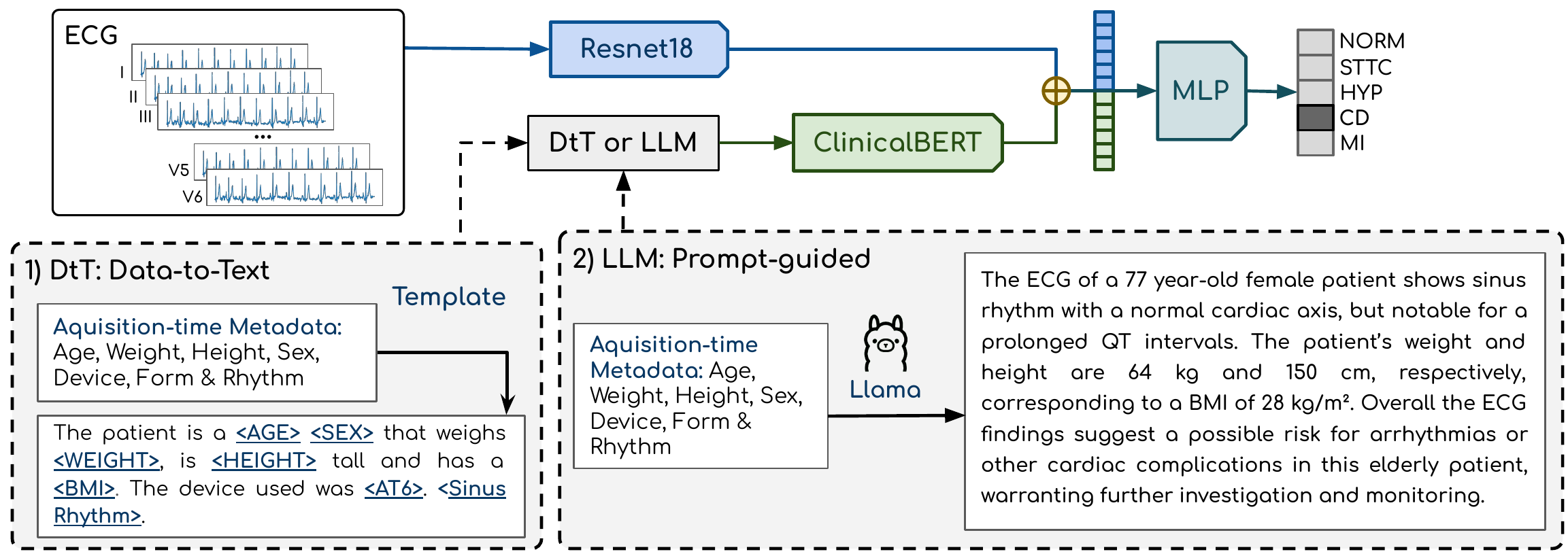}
\end{center}
\caption{Illustration of the \textbf{C}ontextual \textbf{L}anguage-\textbf{I}nformed \textbf{C}ardiac pathology classification (CLIC) framework. The framework consists of two input data workflows: a Resnet18 that receives a 12-lead ECG as input, and a ClinicalBERT that receives a contextual clinical text generated via a template-based strategy called \textbf{Data-to-text} (1) and a \textbf{Prompt-guided} strategy that uses Llama (2).}
\label{fig:clic_workflow}
\end{figure}

In this work, we address this gap in the context of cardiology by proposing \textbf{C}ontextual \textbf{L}anguage-\textbf{I}nformed \textbf{C}ardiac pathology classification (CLIC), a multimodal supervised learning framework that integrates electrocardiogram (ECG) time series signals with textual information for cardiac pathology classification. As depicted in Figure \ref{fig:clic_workflow}, rather than treating text as an auxiliary input, CLIC investigates how contextual language representations can enrich ECG-based feature learning, leading to more informative representations for downstream classification tasks. Specifically, we explore two multimodal configurations of CLIC, which will be detailed further in the next section. These configurations allow us to analyze the impact of increasingly expressive textual context on multimodal representation learning.

Unlike many multimodal approaches in physiological signal analysis that rely on additional clinical sources such as electronic health records (EHRs)~\citep{kim2025multimodal, ecgrepresentationehrlalam2023, yang2021leveragemultimodalehrdata, lyu2023multimodaltransformerfusingclinical}, the proposed CLIC framework operates in a self-contained setting, using only information directly available at acquisition time.

Our results show that incorporating contextual variables improves classification performance and representation quality compared to ECG-only baselines. However, the integration strategy is important. Direct concatenation of numerical attributes provides a strong and stable baseline, yet transforming metadata into natural language yields the best overall results, especially in a deterministic Data-to-Text setting with the ClinicalBERT language encoder. In contrast, LLM-generated reports do not consistently improve the semantic quality of the embeddings beyond structured approaches. Although CLIC is designed to generalize across signal types and datasets, this study focuses on ECG-based diagnosis as a case study to evaluate different strategies for combining physiological signals with structured and text-based contextual information.

\section{Methodology}

We formulate the addressed problem as a supervised multiclass classification task for ECG-based pathology classification. We conduct our experiments using the PTB-XL dataset~\citep{ptbxl_dataset}, a large-scale publicly available ECG benchmark. This dataset meets our computational constraints and provides a rich set of structured metadata, including diagnostic statements and clinical annotations, which support the multimodal experimental setting considered in this work. The classification is performed over five categories: Normal ECG (NORM), Myocardial Infarction (MI), ST/T Change (STTC), Conduction Disturbance (CD), and Hypertrophy (HYP).
We followed the standard split into training, validation, and test sets and observed a notable class imbalance, particularly for Conduction Disturbances and Hypertrophy classes, posing additional challenges for supervised learning. The signals in PTB-XL consist of 500 Hz 12-lead ECGs, accompanied by the patient's data, such as sex, age, and weight, and automatically annotated rhythm and morphology information. We followed the dataset's recommended stratification using the \texttt{strat\_fold} column. Folds 1 through 8 were allocated for training, fold 9 for validation, and fold 10 for testing. The target value corresponds to a single diagnostic class.

As the base model for the ECG, we employ an unimodal ECG classifier based on a ResNet18~\citep{he2015deepresiduallearningimage} architecture adapted for time series modality. The ECG encoder maps each input signal to a 512-dimensional embedding, which is subsequently fed into a multilayer perceptron (MLP) classifier. The MLP consists of two hidden layers with 256, 64, and 5 units, in this order. We employ the same architecture as the backbone in the multimodal scenario.

To incorporate contextual patient information, we evaluate two strategies to construct textual representations from ECG metadata, as shown in Figure \ref{fig:clic_workflow}. These text representations are combined with the ECG signal embeddings. Both strategies use the same metadata variables: age, sex, height, weight, collection device, signal morphology, and rhythm. The height and weight attributes were used to calculate body mass index (BMI), a commonly used metric for weight-related health risks. This selected set of variables corresponds to information that can be collected at acquisition time by health professionals in a clinical environment. The multimodal strategies are:

\textbf{(1) Data to text:} This method converts structured metadata into short natural language descriptions using template-based substitution. It provides deterministic and consistent textual context while preserving clinical reasoning.

\textbf{(2) Prompt guided:} This method generates more complex clinical descriptions using a large language model. In our experiments, we use the 8 billion parameter Llama 3.1~\citep{touvron2023llamaopenefficientfoundation} model to produce a concise clinical ECG report conditioned on available metadata. 

We designed a prompt to guide Llama in generating the ECG reports. The model was instructed to act as a cardiology specialist and produce a single paragraph that simulates a clinical report when given the data described above. To ensure factual consistency, we constrained the model to use only the provided information and to avoid generating false information (e.g., heart rate values not present in the input).

The generated text from both methods were encoded using a pretrained ClinicalBERT~\citep{huang2020clinicalbertmodelingclinicalnotes} model, resulting in a 768-dimensional embedding. The ClinicalBERT parameters are frozen during the model training. Furthermore, we evaluate an additional baseline in which the same metadata used in the multimodal configurations are encoded as numerical features and directly concatenated with the ECG embeddings before classification via the MLP. This baseline was designed to assess whether a language encoder such as BERT can capture extrinsic information beyond what is available through direct numerical feature concatenation. We refer to this model as \textbf{ECG+Attr}.

The models were trained for up to 1000 epochs, with an early stopping patience of 50 epochs. We used a batch size of 16 across all sets, with data shuffling applied exclusively to the training dataloader. The network was optimized using the Adam optimizer with a learning rate of $10^{-3}$, and the loss was computed using Binary Cross-Entropy with unnormalized logits. Finally, to ensure statistical reliability, all experiments were executed across 5 independent runs, and we report the average performance metrics along with their standard deviations.

Both unimodal settings use a two-layer MLP for classification. In multimodal settings, we extend this architecture by adding an MLP layer to aggregate the additional embeddings introduced during the fusion stage. The source code is available at the repository~\footnote{\href{https://github.com/giovanidl/CLIC}{https://github.com/giovanidl/CLIC}}.

\section{Results}
Table \ref{tab:results} summarizes the performance of both unimodal and multimodal architectures across all independent runs. Overall, the results demonstrate that CLIC consistently improves ECG classification when meaningful textual context is incorporated. 

The unimodal ECG baselines achieve strong performance for the majority class (NORM) but struggle with minority and clinically complex conditions, particularly Conduction Disturbance (CD), reflecting the limitations of ECG-only representations. The LLM-generated (CLIC-LLM) multimodal variant shows modest gains for STTC. However, the performance does not consistently surpass the ECG+Attr unimodal baseline across all classes. 

Alternatively, CLIC-DtT achieves the best overall performance, outperforming almost all baseline metrics, including general accuracy and macro-average, while maintaining low variance across runs. Particularly for minority classes, improvements are more pronounced, including CD and STTC, suggesting that structured template-based semantic text provides a more stable integration of demographic and signal metadata than both raw attribute concatenation and LLM-generated descriptions. 
One possible explanation is that LLM-generated reports introduce lexical variability and implicit reformulations that may dilute information. As a result, the textual encoder may extract less stable representations than those from the direct and standardized template sentences.  

For Conduction Disturbance (CD), which is known to be more frequent in older male patients~\citep{haimovich2024frequency, monin2016prevalence}, the inclusion of contextual metadata appears to provide clinically relevant prior information beyond the ECG signal alone. When encoded with a domain-specific language model, the textual representation may incorporate extrinsic medical knowledge about age- and sex-related comorbidities. This additional semantic structure can help the model capture associations that are not explicitly represented in the signal features.

\renewcommand{\arraystretch}{1.2} 
\begin{table}[!h]
\centering
\caption{Performance comparison (mean $\pm$ std) across models using class-wise F1-score, Recall, General Accuracy and Macro-F1. \textbf{Bold} and \underline{underline} values indicate the model with the best and second-best metric for that class, respectively. }
\label{tab:results}
\resizebox{\textwidth}{!}{
\begin{tabular}{lccccccccccc}
\toprule
\multirow{2}{*}{Model} 
& \multicolumn{2}{c}{NORM (1004)} 
& \multicolumn{2}{c}{MI (544)} 
& \multicolumn{2}{c}{STTC (283)} 
& \multicolumn{2}{c}{CD (122)} 
& \multicolumn{2}{c}{HYP (245)} \\
\cmidrule(lr){2-3}
\cmidrule(lr){4-5}
\cmidrule(lr){6-7}
\cmidrule(lr){8-9}
\cmidrule(lr){10-11}
& F1 & Recall 
& F1 & Recall 
& F1 & Recall 
& F1 & Recall 
& F1 & Recall \\
\midrule

ECG Only & 
0.831$\pm$0.009 & 0.862$\pm$0.037 & 
0.705$\pm$0.009 & 0.679$\pm$0.037 & 
0.570$\pm$0.023 & 0.597$\pm$0.025 & 
0.360$\pm$0.013 & 0.308$\pm$0.033 & 
0.553$\pm$0.012 & 0.522$\pm$0.026 
\\

ECG+Attr & 
\underline{0.872$\pm$0.005} & 0.897$\pm$0.023 & 
\underline{0.766$\pm$0.008} & \underline{0.791$\pm$0.013} & 
0.684$\pm$0.021 & 0.701$\pm$0.044 & 
\underline{0.424$\pm$0.017} & 0.334$\pm$0.036 & 
\underline{0.594$\pm$0.016} & \textbf{0.528$\pm$0.038}  
\\

CLIC-DtT & 
\textbf{0.887$\pm$0.010} & \textbf{0.936$\pm$0.018} & 
\textbf{0.781$\pm$0.013} & \textbf{0.792$\pm$0.055} & 
\textbf{0.767$\pm$0.005} & \textbf{0.761$\pm$0.050} & 
\textbf{0.490$\pm$0.051}& \textbf{0.421$\pm$0.069} & 
\textbf{0.597$\pm$0.030} & 0.491$\pm$0.048 
\\

CLIC-LLM & 
0.871$\pm$0.003 & \underline{0.909$\pm$0.011} & 
0.757$\pm$0.009 & 0.750$\pm$0.025 & 
\underline{0.695$\pm$0.009} & \underline{0.712$\pm$0.063} & 
0.401$\pm$0.060 & \underline{0.341$\pm$0.085} & 
0.587$\pm$0.029 & \underline{0.526$\pm$0.082}  
\\

\midrule
\multicolumn{11}{c}{\textbf{Overall Performance}} \\
\midrule

\multicolumn{2}{l}{Metric} 
& \multicolumn{2}{c}{ECG Only}
& \multicolumn{2}{c}{ECG+Attr}
& \multicolumn{2}{c}{CLIC-DtT}
& \multicolumn{2}{c}{CLIC-LLM} \\

\cmidrule(lr){1-2}
\cmidrule(lr){3-11}
\cmidrule(lr){5-6}
\cmidrule(lr){7-8}
\cmidrule(lr){9-10}

Macro-F1
& &\multicolumn{2}{c}{0.604$\pm$0.019}
& \multicolumn{2}{c}{\underline{0.668$\pm$0.006}}
& \multicolumn{2}{c}{\textbf{0.704$\pm$0.015}}
& \multicolumn{2}{c}{0.662$\pm$0.007} \\

General Accuracy
& &\multicolumn{2}{c}{0.714$\pm$0.017}
& \multicolumn{2}{c}{\underline{0.773$\pm$0.005}}
& \multicolumn{2}{c}{\textbf{0.800$\pm$0.008}}
& \multicolumn{2}{c}{0.770$\pm$0.003} \\

\bottomrule
\end{tabular}
}
\end{table}

\begin{figure}[ht]
\begin{center}
\includegraphics[width=0.99\linewidth]{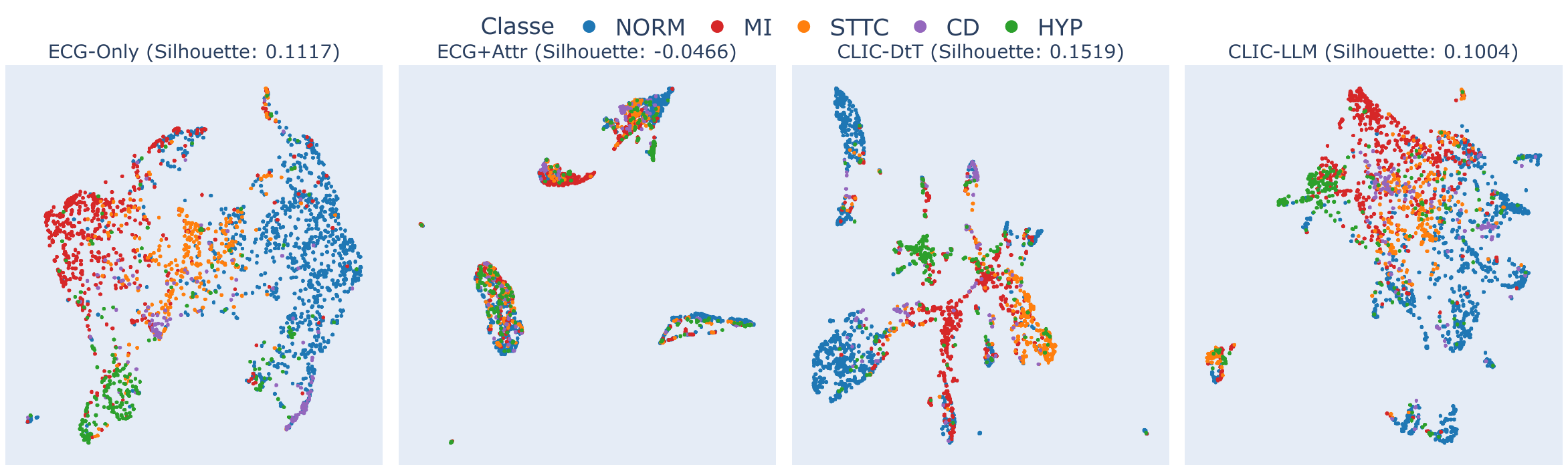}
\end{center}
\caption{UMAP visualization of embedding distributions across models.}
\label{fig:umap_emb}
\end{figure}

To further examine the impact of multimodality on representation learning, we analyzed the embedding space using UMAP~\citep{mcinnes2020umapuniformmanifoldapproximation}, as shown in Figure~\ref{fig:umap_emb}. Analyzing the latent space structure, the ECG-only representations are sparse and unstructured, whereas the ECG+Attr embeddings form dense but highly mixed clusters. The CLIC-DtT model demonstrates superior organization, forming a continuous manifold with more evident separation between normal and pathological classes. This suggests that CLIC-DtT successfully integrates multimodal signals without allowing static attributes to dominate the embedding space.

\section{Conclusion}

Our results show that incorporating contextual patient information improves performance compared to the ECG-only baseline. Even being a simple addition, the direct concatenation of structured attributes ECG+Attr consistently overcomes the ECG-only scenario. Among the multimodal strategies, the Data-to-Text approach achieves greater stability and better overall performance, while the LLM-generated variant yields similar results on some metrics but does not consistently surpass the structured attribute baseline. These findings indicate that deterministic textual generation can provide more stable supervision. In practice, the LLM text did not sufficiently enhance the semantic content of the embeddings to outperform the template-based text in the assessed protocol, suggesting that the additional linguistic complexity did not translate into more discriminative multimodal representations, potentially requiring task-specific alignment or fine-tuning to improve performance. We also observed that contextual metadata is particularly beneficial for the CD class, which is strongly associated with demographic factors. These observations suggest that semantic information captured by the ClinicalBERT encoder boosted the representations' discriminability for classes that are more susceptible to demographic factors.

While this work focuses on a single downstream classification task, it opens directions for future research. These include exploring alternative input signals (e.g., EEG and PPG), novel multimodal fusion strategies, and language model fine-tuning. Overall, CLIC demonstrates that integrating structured and text-based contextual information is a feasible complement to ECG representation learning and it also indicates that deterministic texts may be as effective as fully generative approaches.

\subsubsection*{Acknowledgments}

This study was financed, in part, by the São Paulo Research Foundation (FAPESP), Brazil. Process Number \#2022/03176-1, \#2023/05041-9, and \#2025/13341-8.

\bibliography{iclr2026_conference}
\bibliographystyle{iclr2026_conference}


\end{document}